\documentclass[pdflatex,sn-basic]{sn-jnl}

\jyear{2021}%

\theoremstyle{thmstyleone}%
\theoremstyle{thmstyletwo}%

\theoremstyle{thmstylethree}%

\usepackage{subfigure}

\begin{document}

\title[DIMBA: Discretely Masked Black-Box Attack in Single Object Tracking]{DIMBA: Discretely Masked Black-Box Attack in Single Object Tracking}


\author[1]{\fnm{Xiangyu} \sur{Yin}}

\author[1]{\fnm{Wenjie} \sur{Ruan}}

\author[1]{\fnm{Jonathan} \sur{Fieldsend}}


\affil[1]{\orgdiv{College of Engineering, Mathematics and Physical Sciences}, \orgname{University of Exeter}, \orgaddress{\city{Exeter}, \postcode{EX4 4QF},\country{UK}}
}



\abstract{The adversarial attack can force a CNN-based model to produce an incorrect output by craftily manipulating human-imperceptible input. Exploring such perturbations can help us gain a deeper understanding of the vulnerability of neural networks, and provide robustness to deep learning against miscellaneous adversaries. Despite extensive studies focusing on the robustness of image, audio, and NLP, works on adversarial examples of visual object tracking -- especially in a black-box manner -- are quite lacking. In this paper, we propose a novel adversarial attack method to generate noises for single object tracking under black-box settings, where perturbations are merely added on initial frames of tracking sequences, which is difficult to be noticed from the perspective of a whole video clip. Specifically, we divide our algorithm into three components and exploit reinforcement learning for localizing important frame patches precisely while reducing unnecessary computational queries overhead. Compared to existing techniques, our method requires fewer queries on initialized frames of a video to manipulate competitive or even better attack performance. We test our algorithm in both long-term and short-term datasets, including OTB100, VOT2018, UAV123, and LaSOT. Extensive experiments demonstrate the effectiveness of our method on three mainstream types of trackers: discrimination, Siamese-based, and reinforcement learning-based trackers.}




\maketitle

\section{Introduction}
\label{sec:introduction}
While deep learning has achieved a breakthrough in solving the problems that have been experienced by the artificial intelligence and machine learning community over the past decade, several studies have revealed that Deep Neural Networks (DNNs) are vulnerable to adversarial perturbations~\citep{goodfellow2015explaining} on image processing tasks~\citep{szegedy2014intriguing, moosavidezfooli2016deepfool, xie2017adversarial}. For images, such perturbations are often too small to be perceptible, yet they can completely fool a DNN classifier, detector, or segmentation analyzer, causing them to predict incorrect categories or contours. This leads to great concerns under the circumstances where deep learning models are deployed rapidly in safety and security-critical applications in particular, e.g., self-driving cars, surveillance, drones, and robotics ~\citep{Mnih2015HumanlevelCT}. Besides the computer vision applications, recent works also investigate adversarial attacks on other tasks, \textit{e.g.} natural language processing ~\citep{zhang-etal-2019-generating-fluent}, audio recognition~\citep{audio}, and malware detection~\citep{grosse2017adversarial}.

Single object tracking(SOT), as one of the fundamental problems in computer vision, has recently experienced tremendous improvement through DNNs and plays a significant role in practical security applications such as self-driving systems, robotics, etc.,~\citep{Mnih2015HumanlevelCT}. In terms of the tracking procedure, it can be mainly divided into three categories, Siamese-based trackers~\citep{siamrpn, dasiamrpn, bertinetto2016fullyconvolutional, zhang2019learning}, discrimination trackers ~\citep{prdimp, danelljan2019atom}, and reinforcement learning-based trackers~\citep{yun2017action}. Siamese-based
trackers define the tracking problem as a one-stage detection problem and locate the object that has the most similar feature representation with the initial template on subsequent frames. On the other hand, discrimination trackers predict object locations based on two sub-modules. The first one is target classification, which introduces dedicated optimization techniques to discriminate between the background and the target object, then the target estimation module is exploited to regress an intersection-over-union (IoU) score between the ground-truths and predicted bounding boxes. The third category, reinforcement learning-based trackers, formulates the whole tracking procedure as a Markov Decision Process, and selects different actions according to the agent state at the current step. However, after the concept of adversarial attack was proposed by~\citep{szegedy2014intriguing}, although intensive follow-up methods were inspired to demonstrate various adversaries to deceive deep learning models~\citep{goodfellow2015explaining, kurakin2017adversarial, madry2019deep}, adversarial robustness concerning object trackers has yet been fully explored. As far as we know, only a handful of research works appeared very recently. For example, some researchers \citep{yan2020cooling} have proposed a Cooling-Shrinking Loss to train the perturbation generator to achieve an effective and efficient adversarial attacking algorithm. Moreover, spatial-temporal sparse noise was applied in~\citep{guo2020spark} along targeted or untargeted trajectories. By categorizing the tracking problem into classification and regression branches, researchers in \citep{chen2020one} focused on free-model object tracking with dual attention. 
\begin{figure}[htbp]
\centering
\includegraphics[width=1.0\textwidth, height=0.33\textwidth]{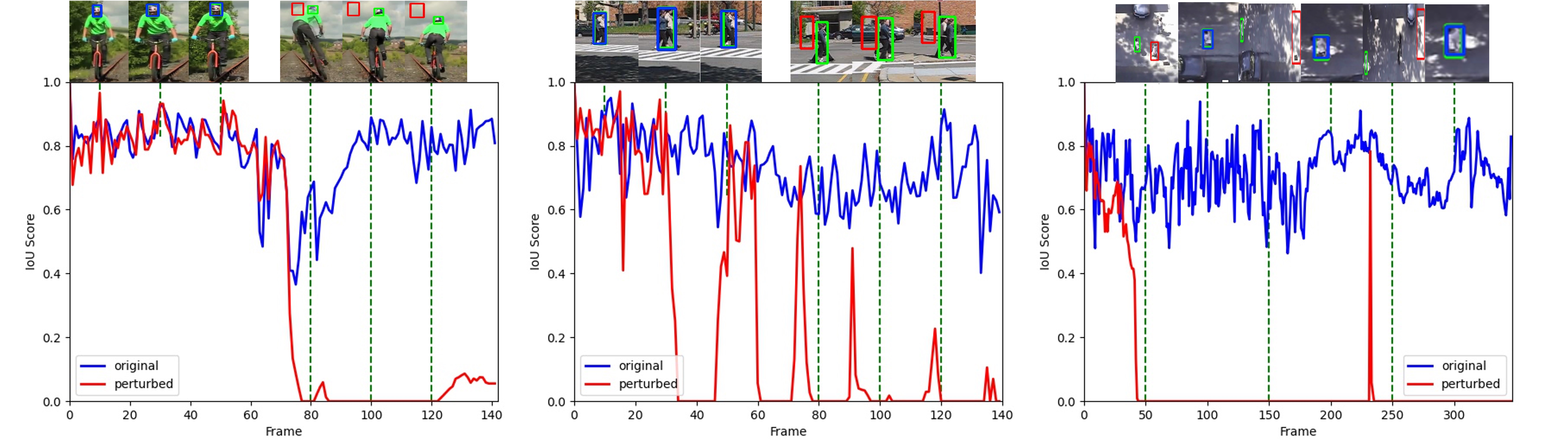} 
\caption{Visualization of tracking results generated by trackers from three different tracking categories under DIMBA Attack, including SiamRPN++~\citep{siamrpn++}(left), ADNet~\citep{yun2017action}(middle), and PrDiMP~\citep{prdimp}(right). Clipped frames above the chart qualitatively demonstrate the behaviors of trackers with or without attack. Green bounding boxes refer to ground truths, blue ones measure original tracking results, and red ones illustrate failed tracking performance. The charts below indicate IoU scores between predicted bounding boxes and ground truths, and the tracking performance with or without attack is separately represented in red and blue lines.}
\label{fig1}
\end{figure}

Whereas current attacking techniques applied on SOT exhibit several limitations that may severely restrict their generality in practice. Specifically, we highlight the following disadvantages: (1)\textit{Most tracking adversaries cannot be extended to constrained black-box SOT applications}. Given comprehensive knowledge of model architecture and parameters, miscellaneous approaches are capable of generating effective perturbations over the whole video clip based on the computation of network gradient. However, the target network is often inaccessible within safety-critical scenarios where we can only obtain hard-label predictions during the whole tracking procedure. Therefore, practical black-box attack algorithms are worthy of exploration. (2)\textit{Current methods compose perturbations often on multiple frames.} As illustrated above, existing white-box attacks can realize powerful overall results, but most of them are derived from noises attached to a large portion of frames. Although the initial frame of a video plays a vital role in SOT, few works pay attention to this, either in white-box or black-box scenarios. For instance, the Hijacking algorithm~\citep{yan2020hijacking} generates an adversary on a special clip of the video, and the IoU attack~\citep{jia2021iou} proposes a continuous black-box attack framework imposed from the $2_{nd}$ frame to $N_{th}$ frame. (3)\textit{Recent black-box attack algorithms applied on SOT do not consider computational efficiency}. As far as we know, none of the existing black-box attacks on SOT considers query efficiency. ~\citep{liang2020efficient} presents a transferable attack mode, but it is specialized in white-box cases.~\citep{jia2021iou} focuses on temporal correlations between adjacent frames, but its effectiveness heavily relies on the length of a video and the query times per frame. 

Different from the attack on image classification or segmentation tasks where perturbation can be merely added on a single picture, the evaluation metrics in SOT are determined by the whole video clip. As the number of perturbed frames increases, adversaries will be detected more easily. Meanwhile, the knowledge of video gradient is completely lacking within a black-box scenario. Therefore, a sacrifice of query times is almost unavoidable to improve adversarial results. Subsequently, we propose a question: 
\vspace{1.0mm}
\begin{center}
    \textit{Can we combine efficiency and effectiveness in black-box attack on SOT}?
\end{center}
\vspace{1.0mm}
Or in other words, can we select the most fragile part of a video to perturb and reach destroyed tracking results more quickly? In this paper, we combine query-based method with reinforcement learning framework and propose the \textit{\underline{Di}screte \underline{M}asked  \underline{B}lack-Box} \underline{a}ttack(DIMBA) algorithm on SOT, where we simply modify initialized frames across the whole video clip to realize perturbed results. In contrast to previous works, we reversely craft heavy or effective perturbations at first, then decrease the adversarial magnitude using a modified sign attack method. In summary, the key contributions of our paper are as follows:

 \begin{figure}
\centering
\includegraphics[width=\textwidth]{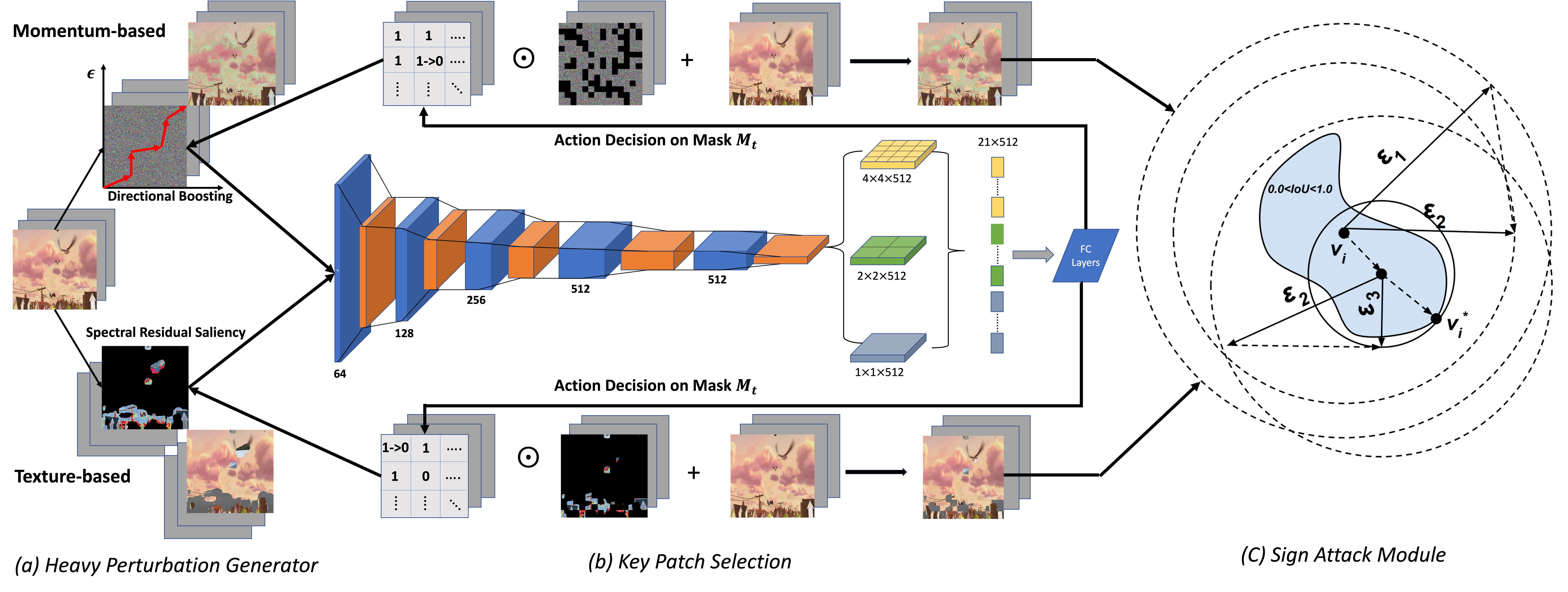}
\caption{Overview of DIMBA framework, which contains heavy perturbation generator, key patch selection, and sign attack module, (a) Heavy Perturbation Generator initially constructs candidate adversarial videos, originating from either momentum-based approach or texture-based approach. Partial adversaries are overly perturbed, which are therefore sent to subsequent components. (b)Then, Key Patch Selection assigns the mask value of particularly perturbed patches to 0 based on an Actor-Critic network, of which structure is proposed above. (c) Sign Attack Module estimates gradients around designated directions optimized from previous steps and computes final results.}
\label{fig2}
\end{figure}

\begin{itemize}
\item[\textbf{1)}]We formulate the black-box attack problem on SOT in a more practical and query-efficient manner. Compared to recursively generating perturbed results in each frame, we focus on initialized frames, which boosts the attack efficiency.

\item[\textbf{2)}]To reduce unnecessary perturbations with large adversarial magnitude on specific areas in initialized frames, and increase the probability of generating perturbations causing similar attack performance within a smaller perturbing radius, we introduce an A2C (Actor-Critic) grid searching strategy 

\item[\textbf{3)}]The comprehensively devised experiments over OTB100, UAV123, LaSOT, and VOT2018 datasets show that DIMBA attack can generate imperceptible perturbations more efficiently, and achieve competitive or even better performance compared to SOTA black-box attacks on SOT.
\end{itemize}

\section{Related Work}
\label{related}
\subsection{Adversarial Attacks on Visual Object Tracking}
Wide applications of visual object tracking have led to numerous specialized real-world techniques, which have also resulted in well-crafted attacks from the adversarial perspective. Taking the realm of physical world attacks into account, ~\citep{eykholt2018robust} analyzed adversarial stickers on stop signs in the context of autonomous driving to fool YOLO ~\citep{redmon2016look}. ~\citep{jia2019fooling} proposed a `tracking hijacking' technique to fool multiple object trackers with imperceptible perturbations computed for object detectors in the perceptual pipeline of autonomous driving. Meanwhile, ~\citep{yan2020cooling} developed an attacking technique to deceive single object trackers based on SiamRPN++ ~\citep{siamrpn}. Their method trains a generator model to construct adversarial frames under a `cooling-shrinking' loss, which is manipulated to cool down the hot target regions and force the bounding boxes to shrink during online tracking. \citep{huang2020universal} delved into physical attacks on object detectors in the wild by developing a universal camouflage for object categories. A one-shot adversarial attack is demonstrated in~\citep{chen2020one} for single object tracking were inserting a patch in the first frame of the video results in losing the target in the subsequent frames. A spatial-aware attack (SPARK) is proposed in~\citep{guo2020spark} fool online trackers. This approach imposes an $L_{p}$ constraint over perturbations while computing them incrementally based on previous frames. Extensive experiments show that their adversaries are capable of fooling multiple state-of-the-art trackers. 

Differing from previous attacking models in white-box settings,~\citep{jia2021iou} explores black-box perturbations by making use of temporally correlated information and incrementally adding noise from the initial frame to subsequent frames. However, it focuses extensively on locally anchored noise between adjacent templates and is devoid of long-term diversity.

\subsection{Deep Reinforcement Learning}
Due to its ability to scale to previously intractable decision-making problems, Deep Reinforcement Learning (DRL) has been a growing area recently. Kickstarting this revolution~\citep{Mnih2015HumanlevelCT}, for example, firstly learns to play a range of Atari 2600 video games at a superhuman level directly from pixel-level knowledge, whilst demonstrating that RL agents could be trained on raw, high-dimensional observations based on reward signals. As another standout success, AlphaGo~\citep{silver2016mastering} parallelled the historic achievement of IBM's Deep Blue and defeated a human world champion in Go.

\section{Methodology}
\label{method}
In this section, we first introduce the preliminaries of our proposed attack method. The details of DIMBA are presented in subsequent sections. The general pipeline of our algorithm is shown in \ref{fig2}. Initialized frames are taken as an input(For simplicity, we only consider One Pass Evaluation(OPE)in the following parts). With a momentum-based perturbation, generator simulating the optimal gradient descent direction and exploiting historical noise trajectory as shown in MI-FGSM~\citep{dong2018boosting}, and a texture-based approach selecting candidates by crafting spectral residual detection, we accumulate bunches of candidate first frames. Then an Actor-Critic agent computes the importance of patches segmented equally in the initial frame and selects the least important region under the current state. Last but not least, an iterative boundary-walking strategy is utilized to compress perturbation magnitude while maintaining attack results within a specific region.

\subsection{Preliminaries}
We denote a video sample by $v \in\mathcal{V}\subset{\mathbb{R}^{N\times H\times W\times C}}$ with \textit{N, H, W, C} referring to the number of frames, height, width, and the number of channels respectively. A specific frame can be denoted as $v_{i}(i\in {1,...N})$, where $N$ is the length of video $v$. Generally, SOT learns a tracking model $\mathcal{T}(v;\theta) : \mathcal{V} \rightarrow \mathcal{(B, S)}$ by minimizing regression loss between ground truth and predicted bounding boxes in each frame and maximizing similarity of predicted bounding boxes between adjacent frames. $\mathcal{B} \in \mathcal{R}^{N\times 4}$ indicates localizing matrix, where each row $[x_{i}, y_{i}, w_{i}, h_{i}]$ denotes the x-axis and y-axis coordinates, width, and height of the predicted bounding box for $v_i$. Meanwhile, $\mathcal{S}$ collects the highest confidence scores for each frame. According to the evaluation method, SOT can be summarized into two categories. The first one initializes only once in a single video, which is also called One Pass Evaluation (OPE). In contrast, the second approach can restart the tracker several frames after the failed one, such as testing trackers on Visual Object Tracking Challenge 2018~\citep{vot2018}. The goal of an adversarial attack in SOT is to find an adversarial example $v^{*}$ that can fool the network to make a shifted or even target-lost bounding box in the sequence, while keeping $v^{*}$ within the $\epsilon$-ball centered at $v$ using $L_p$ normalization $\|v^{*} - v\|_{p}$, where p can be 1, 2 or $\infty$. Here in this paper, we mainly focus on the $L_\infty$ norm and SSIM similarity ~\citep{ssim} for comparison to clean frames.\par
Although there are multiple evaluation metrics for SOT across various challenges, we decide to explore two standards that are in most common use for visual tracking, represented as $\mathcal{A}$ and $\mathcal{R}$, short for accuracy and robustness. $\mathcal{A}$ denotes the average of $IoU$ scores of all frames that contain overlapping perturbed bounding boxes and predicted bounding boxes until the end of video or reinitialization. $\mathcal{R}$ then weights the tracking performance according to the number of failed frames in a discounted reward manner. These two values can be calculated as:
\begin{equation}
IoU_i = \frac{\hat{B}_i \cap B_i}{\hat{B}_i \cup B_i},\quad
ro_{i} =\left\{
\begin{aligned}
1 & , & IoU_i \in(0,1], \\
0 & , & else.
\end{aligned}
\right.
\end{equation}

\begin{equation}
\mathcal{A} = \frac{1}{N}\times\sum_{i}^{N} (\gamma_{a})^{i//L} IoU_i*ro_i,\quad \mathcal{R} = \sum_{i}^{N}(\gamma_{r})^{i//L}ro_{i}
\end{equation}
where $IoU_i$ represents \textit{Intersection over Union} between predicted $\hat{B}_i$ and ground truth $\hat{B}_i$. $\gamma_{a}$ and $\gamma_{r}$ state the discounted factors for accuracy and robustness, highlighting the impact of future tracking performance. Generally in our work, both of them are set to 0.9. Similar to SPARK ~\citep{guo2020spark}, we split the video into $L$-length intervals based on a common frame rate (also known as Frame Per Second ($FPS$)), considering weight factors within the same interval are supposed to be set equivalently, but decreased exponentially in a long term view. Generally, attacks on SOT can be categorized into untargeted and targeted attacks. An untargeted attack generates an adversarial example either from a long-term or short-term tracking perspective according to object motion, aiming to decrease the average value of $IoU_i$ in a whole video clip, which in the best case can cause the tracker to lose the target. In contrast, a targeted attack focuses on the object trajectory or shape of the bounding box. In this paper, we will mainly focus on untargeted attacks.

\subsection{Heavy Perturbation Generator} In the first stage of our proposed pipeline, we generate a heavily perturbed initial frame. We synergistically exploit texture-based and momentum-based generators to produce adversarial candidates to diversify adversarial directions and increase the probability of successful perturbations. Take \textbf{texture-based} perturbations, for instance, we randomly select a certain number of videos from the current dataset and pick up frames from candidates with the same timestamp as the victim frame. \begin{algorithm}[!ht]
  \caption{Momentum-based perturbation generation in OPE}
  \label{momentum}
  \begin{algorithmic}[1] 
     \Require SOT tracker $\mathcal{T}$, clean video $v$, adversarial video $v^*=v$, maximum perturbation $\epsilon$, candidate number $C$, momentum factor $\mu$, trade-off factor $\iota$, iterations $k$, initial gradient $g_{0}$, tracking performance $\mathcal{TP}=1$, adversarial candidate set $\mathcal{V}$.
     \Ensure  adversarial candidate set $\mathcal{V}$
     \State $\mathcal{A}$, $\mathcal{R}$ = \textit{$\mathcal{T}(v;\theta)$}, \; $\mathcal{R^*}= R$, \;
     \While{$\mathcal{R^*} \leq \mathcal{R}\;\textbf{or}\; \|v_0^*-v_0\|_{\infty} \leq \epsilon$}
     \For{$i=0$ to $C-1$}
     \State $v_{0}^{'} = v_{0}^* + \mathcal{N}(0, I, v_{0}^*.shape);\quad g^{'} = \frac{v_{0}^{'}-v_{0}^*}{\|v_{0}^{'}-v_{0}^*\|_{\infty}}$
     \State $\mathcal{A}^*, \mathcal{R}^* = \mathcal{T}(v_{0}^{'}; \theta)$
     \If{$\iota\times\frac{\mathcal{A}^*}{\mathcal{A}}+(1-\iota)\times\frac{(\mathcal{R}+1)}{(\mathcal{R}^*+1)}< \mathcal{TP}$}
     \State $g_{opt} = \mu\times g_{0} + g^{'};
     \mathcal{TP}=\iota\times\frac{\mathcal{A}^*}{\mathcal{A}}+(1-\iota)\times\frac{(\mathcal{R}+1)}{(\mathcal{R}^*+1)}$
     \EndIf
     \EndFor
     \State $v_{0}^* = v_{0}^* + \frac{\epsilon}{k}\times Sign(g_{opt});\;\;g_0=g_{opt};\;\;\mathcal{V}.append(v_0^*)$
     \EndWhile
     \State \textbf{Return} $\mathcal{V}$
  \end{algorithmic}
 \end{algorithm}
Particularly in OPE scenarios, the victim frame would be \#0. Then considering both human visual systems and video processing models that concentrate on target locations contributing more to final results, we apply a Spectral Residual Saliency approach~\citep{hou2007saliency} on each candidate using pixel-wise mask $M^p\in\{0,1\}^{S\times W\times H\times C}$, where $S, W, H, C$ indicate the number of reinitialization($S=1$ for OPE), width, height, and the number of channels for each video. Then all candidates will be appended to adversarial sets $\mathcal{V}$.  \textbf{Momentum-based} approach, on the other hand, is a technique for accelerating gradient descent algorithm by accumulating a velocity vector in the gradient direction. IoU-Attack~\citep{jia2021iou} leverages this concept and extends it to temporal correspondence among continuous frames. Inspired by these works, we present a novel spatial momentum-based approach, which is applied to the initial as well as the most essential frame of a video. As illustrated in Algorithm \ref{momentum}, by randomly sampling perturbing directions in each attack level denoted as $\frac{\epsilon}{k}$, where $\epsilon$ indicates the magnitude of $L_\infty$ normalization, we craft adversaries along the historically optimal direction progressively, until we find the successful perturbation on the initial frame or the magnitude of perturbation exceeds the $\epsilon$-ball bound around $v_i$. Balanced by trade-off factor $\iota$, if the tracking performance decreases, we then update and get the optimal gradient $g_{opt}$ with momentum. With two different perturbation generators, we can finally obtain an adversarial set $\mathcal{V}$ full of heavily destroyed initial frames. Then we feed them into the next part of our pipeline. For simplicity, only the OPE-based case is summarized in Algorithm 1. Cases with reinitialization (VOT2018) can be easily extended by repeating the previous process on all reinitialised frames step by step.

\subsection{Actor-Critic Key Patch Selection}
As illustrated above, some areas in the initial frame are more beneficial for feature representations of the target object, but others are not. Take video \textit{Bird1} in Figure \ref{fig2} for instance, perturbations added to corners affect much less than those on more significant regions, like bird-surrounding ones. Therefore, removing redundant perturbations attached to those regions will not affect the whole attack results (or at most only marginally) but decrease adversarial magnitude for perturbations. As shown in Figure \ref{fig2}, we impose a mask that is split into $\mathcal{P}\times\mathcal{P}$ patches and element-wisely composed of all 1s. Considering computational efficiency as well as the averaged size of video frames across different datasets, we adjust $\mathcal{P}$ as a hyper-parameter and conduct a grid search. Then we apply a reinforcement learning (RL)-based key patch selection framework, which is implemented by 
\begin{figure}[htbp]
\centering
\includegraphics[width=1.0\textwidth]{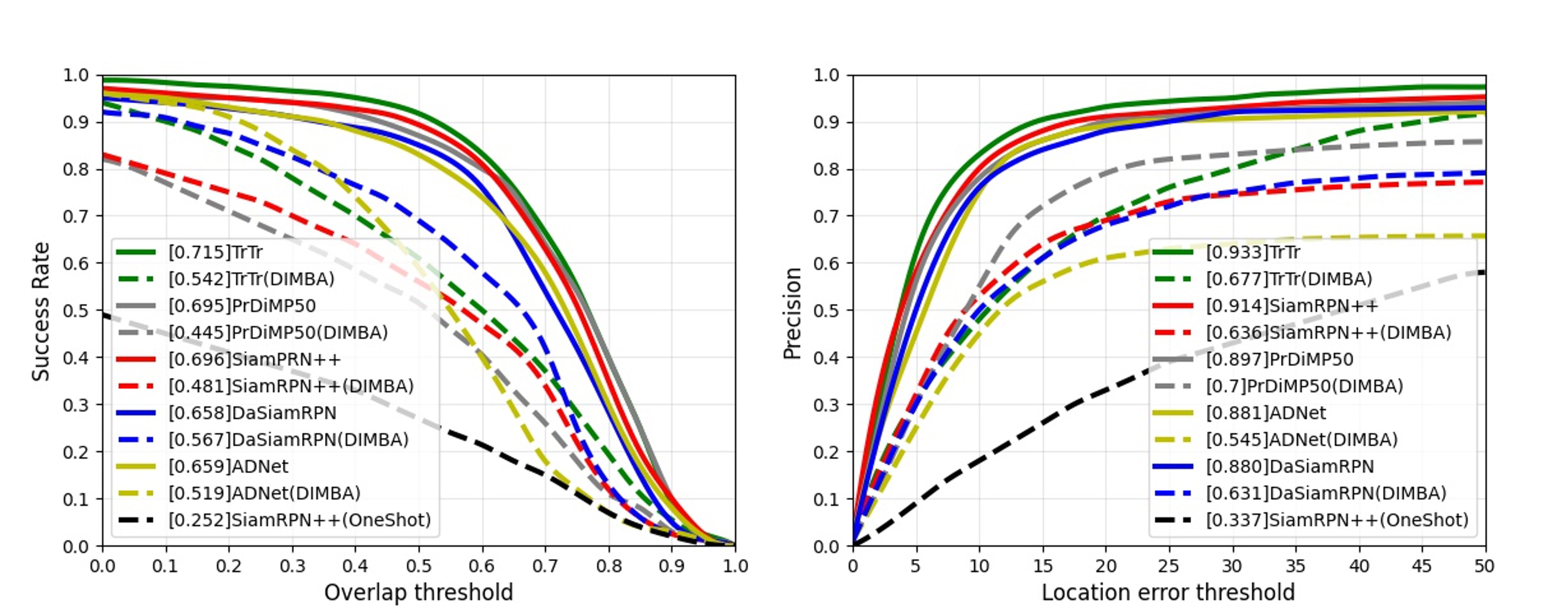}
\caption{Success and Precision Plots of trackers with or without adversarial attacks on OTB100 dataset}
\label{otb100_img}
\end{figure}
Actor-Critic network $\mathcal{Z}$, to select the least important patch step by step until the RL agent enters into a terminal state.\par
As shown in the second part of Figure \ref{fig2}, our network contains 5 convolutional layers, each of them is followed by a max-pooling layer, where parameters are shared between Actor and Critic branches, and extract features of newly added perturbations. However, the shape of videos can be varied even in the same tracking dataset. Resizing them into a fixed size may result in unwanted geometric distortion, which is extremely harmful to localizing objects in SOT. Therefore we introduce a Spatial Pyramid Pooling (SPP)~\citep{He_2016_CVPR} strategy on top of the last convolutional layer to remove the fixed size constraint of the network. Subsequently, we append 3 fully connected layers to estimate what is the best action that the agent should take and the corresponding critic value of that. \par
Generally, we consider the key patch selection as a multi-step Markov Decision Process (MDP), which contains states, actions, transition function, and a reward function. In our task, the state $s_{t}$ at time step $t$ is defined as the pixel-wise difference between $v_{0}$ and $v_{0}^*$ masked by the current mask $M_t\in\mathbb{R}^{S\times\mathcal{P}\times\mathcal{P}}$. It can be denoted as:
\begin{equation}
    s_{t} = (v_{0}^* - v_{0})\odot M_{t}
\end{equation} where $\odot$ represents Hadamard product. At time step 0, $M_0$ is $\{1\}^{S\times\mathcal{P}\times\mathcal{P}}$. An action $a_t = \mathcal{Z}(s_t)$ refers to a $S\times \mathcal{P}^2$ softmax matrix, indicating the least important patch in each initialized frame to successfully track the target at time step $t$. Then once the agent chooses an action $a_{t}$, we can set the corresponding element in $M_t$ to 0.
\begin{algorithm}[t]
\caption{Key Patch Selection and Sign Attack Module in OPE}
\label{alg2}
\begin{algorithmic}[1]
\Require SOT tracker $\mathcal{T}$, clean video clip $v$, A2C pretrained policy $\theta_p$,
value network parameter $\theta_c$, adversarial candidate set $\mathcal{V}$, video candidate number $n$, gradient candidate number $K$, smoothing parameter $\rho_d$, and direction learning step size $\alpha$. number of attack queries $\mathcal{N_A}$, initial grid mask $M$
\Ensure adversarial example set $\mathcal{V}$
\State Fine-tune A2C network parameters $\theta_p$ and $\theta_c$ using top-$n$ adversarial videos from $\mathcal{V}$ that is Ranked based on $\mathcal{TP}$ in Algorithm 1 and $\|\mathcal{V}_i-v_0\|_{\infty}$ with ascending order.
\For{$i=0$ to $n$}
\State $\mathcal{A}, \mathcal{R} = \mathcal{T}(\mathcal{V}_i;\theta)$
\State Apply policy $\theta_p$ to get sparse mask $M_i$
\State $\mathcal{A^*}, \mathcal{R^*}$=$\mathcal{T}((\mathcal{V}_i-v_0)\odot M_i+v_0; \theta)$
\If{$\gamma\frac{\mathcal{A^*}}{\mathcal{A}}+(1-\gamma)\frac{\mathcal{R}}{\mathcal{R^*}}\leq\kappa$}
\State $\phi_d=\frac{(\mathcal{V}_i-v_0)}{\|\mathcal{V}_i-v_0\|_{\infty}}$
\State Using binary search algorithm to compute $g(\phi_d)$ with $\kappa=\frac{\gamma(\tau_1\tau_2-1)+1}{\tau_2}$
\EndIf
\For{$n_A=0$ to $\mathcal{N_A}$}
\State Randomly sample $K$ vectors $u_1,...,u_k$ using Gaussian distribution $\mathcal{N}(0, I)$
\State $\hat{\nabla}g(\phi_d) = \frac{1}{K}\sum_{k=1}^{K}Sign(g(\phi_d+\rho_d u_k)-g(\phi_d))u_k$
\State $\phi_d=\phi_d-\alpha\hat{\nabla}g(\phi_d)$
\State Recompute $g(\theta_d)$ as shown above.
\EndFor
\State $\mathcal{V}_i=v_0+g(\phi_d)\phi_d$
\EndFor
\State Ranking $\mathcal{V}$
\State \textbf{Return} $\mathcal{V}$

\end{algorithmic}
\end{algorithm}
Denoting this process as a function $\mathcal{F}$, we can update the state to
\begin{equation}
    s_{t+1} = (v_{0}^* - v_{0})\odot \mathcal{F}(M_{t}, a_{t}) 
\end{equation}

$s_{t+1}$ will be the terminal state if $a_{t} \in \{a_{0}, a_{1}, ..., a_{t-1}\}$ or $\frac{\mathcal{A}(\mathcal{T}(v_{0}+s_{t+1}))}{\mathcal{A}(\mathcal{T}(v_{0}+s_{0}))} > \tau_1$ or $\frac{\mathcal{R}(\mathcal{T}(v_{0}+s_{t+1}))}{\mathcal{R}(\mathcal{T}(v_{0}+s_{0}))} < \tau_2$. Since SOT is inherently a regression problem within the continuous output space instead of a pure classification problem, slight manipulation of the adversarial perturbation may be reflected in the final tracking results. Therefore we introduce ratio thresholds $\tau_1$ and $\tau_2$ to maintain the attack results within an acceptable scale. Generally, our goal is to delete less important patches and maximize the long-term expected reward, therefore we design the reward in step $t$ as 
\[ r_t = 
             \begin{cases}
             0,\quad\quad\quad\quad\quad\quad\quad\quad\quad\quad\quad a_{t} \in \{a_{0}, a_{1},.., a_{t-1}\};\\
             -1,\quad\quad\frac{\mathcal{A}(\mathcal{T}(v+s_{t+1}))}{\mathcal{A}(\mathcal{T}(v+s_{I}))} > \tau_1\;or\;\frac{\mathcal{R}(\mathcal{T}(v+s_{t+1}))}{\mathcal{R}(\mathcal{T}(v+s_{I}))} < \tau_2;\\
             \gamma\frac{\mathcal{A}(\mathcal{T}(v+s_{I}))}{\mathcal{A}(\mathcal{T}(v+s_{t+1}))}+(1-\gamma)\frac{\mathcal{R}(\mathcal{T}(v+s_{t+1}))}{\mathcal{R}(\mathcal{T}(v+s_{I}))},\quad\quad\quad else
             \end{cases}
\]
In the offline training stage, we select a certain number of candidate videos generated from the previous step, then feed them into policy network $\pi_{\theta_c}(a_t\|s_t)$ and critic network $\pi_{\theta_c}(c_t\|s_t)$ to maximize the expected long-term reward with 
\begin{figure}[htbp]
\centering
\includegraphics[width=1.0\textwidth]{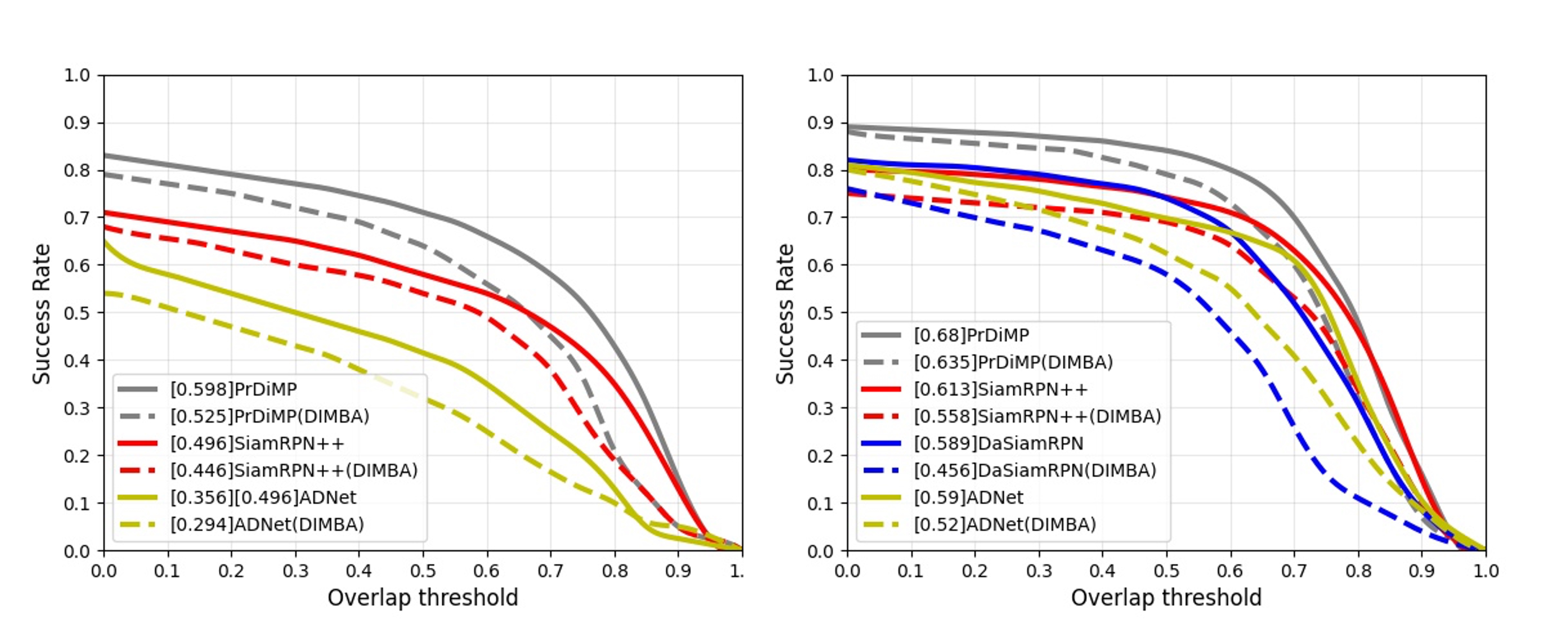}
\caption{Success Plots of trackers with or without adversarial attacks on UAV123 and LaSOT}
\label{uavlasot}
\end{figure}
PPO algorithm, which is written as
\begin{small}
\begin{eqnarray}
\label{eq}
L(\theta_{p})=\sum_{(s_t, a_t)}\min\left(\frac{\pi_{\theta_p}(a_t\|s_t)}{\pi_{\theta_p^{old}}(a_t\|s_t)}, clip\left( \frac{\pi_{\theta_p}(a_t\|s_t)}{\pi_{\theta_p^{old}}(a_t\|s_t)}, 1-\rho, 1+\rho \right)\right)A_{\theta^{old}_p}(s_t\|a_t)
\end{eqnarray}
\end{small}
where $A_{\theta_p}(s_t\|a_t) = Q_{\theta_p}(s_t, a_t)-V_{\theta_c}(s_t)=\gamma^{T-t}V(s_T)+\gamma^{T-t-1}r_{T-1}+\dots+r_t-V_{\theta_c}(s_t)$, $Q_{\theta_p}$ is the Q-value calculated by discounting future rewards, $V_{\theta_c}$ is the critic value generated by critic network. $\rho$ denotes the clip parameter to regularize policy iterations.\par

\subsection{Sign Attack Module}
As indicated in Algorithm \ref{alg2}, after removing less important patch-level perturbations attached to initial frames of videos, we can fetch manipulated adversarial examples as well as their tracking accuracy and robustness. Then we need a boundary walking method to help us compress the noise magnitude while maintaining attack results within a specific scope.
As shown in part (c) of Figure \ref{fig2}, we iteratively update victim frame $v_0$ until its magnitude is compressed from $\epsilon_1$ to $\epsilon_3$, while maintaining competitive attack results or even strengthening it. Cheng \textit{et al.}~\citep{cheng2018query} states that a black-box attack problem can be formulated into an optimization problem, where the objective function can be evaluated as a binary search with additional model queries. Then a zeroth-order optimization algorithm can be applied to solve this optimization problem. In this paper, we exploit the Sign-OPT algorithm in the Sign Attack Module.\par
In our approach, $\phi_d$ and $g(\phi_d)$ indicate our designated search direction and corresponding distance from the initial frame $v_{0}$ to its nearest adversarial example that has the same or similar tracking results within a predefined threshold along $\phi_d$. The objective function can be written as
\begin{small}
\begin{equation}
\underset{\phi_d}{\min}\;g(\phi_d),\quad where\;g(\phi_d)=\underset{\lambda}{\arg min}(\mathcal{AR}(\mathcal{T}(v_0+\lambda\frac{\phi_d}{\|\phi_d\|};\theta))\leq \kappa)
\end{equation}
\end{small}
\begin{figure}[htbp]
\centering
\includegraphics[width=1.0\textwidth]{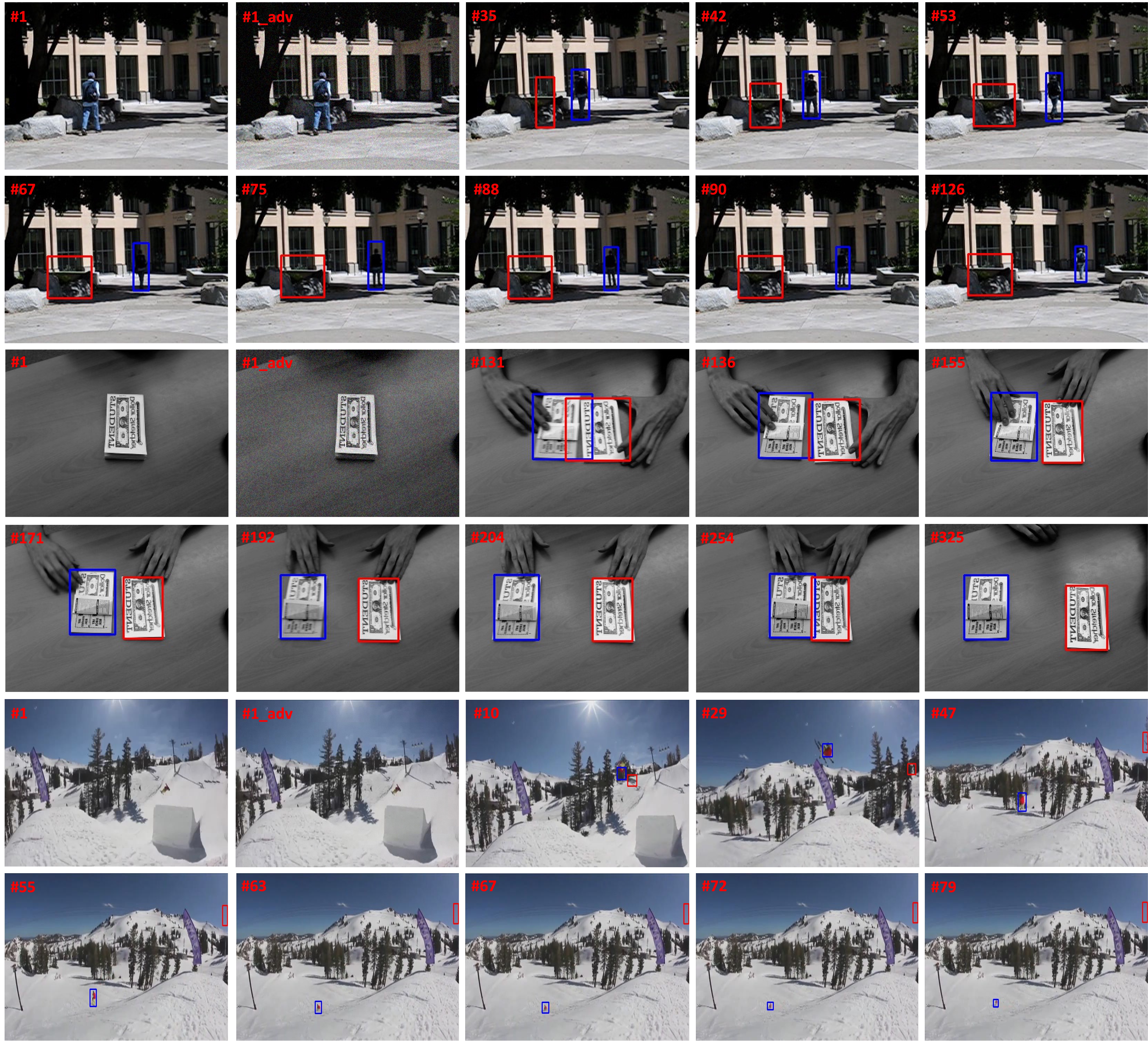} 
\caption{Illustration of clean and adversarial tracking results collected from DIMBA attack on PrDiMP50 tracker. Blue Bounding boxes indicate originally predicted bounding locations while red ones demonstrate attacked ones.}
\label{qualitative}
\end{figure}
which can be evaluated by a local binary search procedure. As the evaluation results of SOT, $\mathcal{AR}$ is denoted as $ \gamma\frac{\mathcal{A}(\mathcal{T}(v_0+\lambda\frac{\phi_d}{\|\phi_d\|}))}{\mathcal{A}(\mathcal{T}(v_0+s_0))}+(1-\gamma)\frac{\mathcal{R}(\mathcal{T}(v_0+s_0))}{\mathcal{R}(\mathcal{T}(v_0+\lambda\frac{\phi_d}{\|\phi_d\|}))}$. We need to estimate its directional derivative by consuming a huge amount of queries when computing $g(\phi_d +u)-g(\phi_d)$. However, it will take a large number of computational resources if we intend to obtain the gradient derivative accurately. Due to the various and large dimensions of our input, we decide to improve query complexity by an imperfect but informative estimation of directional derivative. Therefore, we exploit the sign value and compute the gradient by sampling $K$ gaussian vectors:
\begin{small}
\begin{equation}
\hat{\nabla}g(\phi_d) = \frac{1}{K}\sum_{k=1}^{K}Sign(g(\phi_d+\rho_d u_k)-g(\phi_d))u_k
\end{equation}
\end{small}
When starting an attack on videos, we need to initialize perturbing directions $\phi_d=\frac{v_0^*-v_0}{\|v_0^*-v_0\|}$, where $v_0^*$
can be retrieved by sampling from $v_0$'s candidate adversarial sets $\mathcal{V}$, 
including texture-based and momentum-based perturbations. Detailed in Algorithm 2, by trading off the magnitude of adversaries and their tracking performance, we rank the candidate list with $\mathcal{TP}$ and $L_1$ normalization and pick the top-$n$ target video clips for the attacked video.

\section{Experiments}
In this section, we describe our experimental settings and analyze the effectiveness of the proposed DIMBA algorithm against different trackers on four challenging short-term or long-term datasets, including OTB100~\citep{otb}, VOT2018~\citep{vot2018}, UAV123~\citep{uav123}, and LaSOT~\citep{fan2019lasot}. Part of the qualitative tracking results performed by PrDiMP50 is shown in Figure \ref{qualitative}

\subsection{Experimental Settings}
\textbf{Victim Models}. As mentioned in section 1, current tracking models can be divided into Siamese-based, discrimination, and reinforcement learning-based trackers. 
\begin{figure}
\centering
\includegraphics[width=\textwidth]{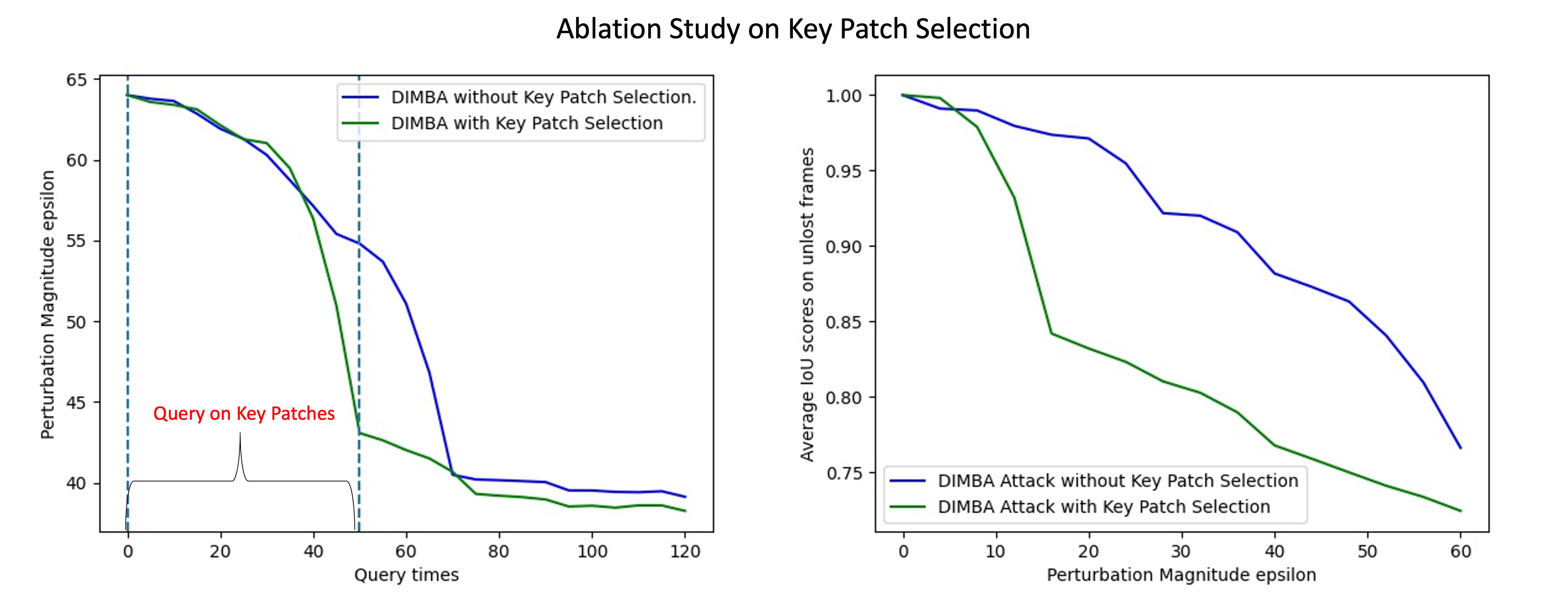}
\caption{Illustration of the ablation study on key patch selection module of our proposed DIMBA Attack. Results are averaged over the OTB100 dataset tracked by PrDiMP50. The left figure indicates the fluctuation of perturbation magnitude with respect to query times. While the right one denotes the relation between the average overlap score on each frame and perturbation magnitude.}
\label{fig3}
\end{figure}
Considering overall tracking performance, we select one or more most representative trackers for each of them, which consists of SiamRPN++ that uses AlexNet~\citep{alexnet}, mobilenetv2 ~\citep{sandler2018mobilenetv2}, and ResNet50 ~\citep{He_2016_CVPR} as backbones, DaSiamRPN~\citep{dasiamrpn}, PrDiMP~\citep{prdimp}, and Action-Decision Network~\citep{yun2017action}.\\
\textbf{Metrics}. To fairly compare our attack results with original tracking performance and previous black-box attacks on SOT, standard evaluation methods are exploited. While testing DIMBA on OTB100~\citep{otb}, UAV123 ~\citep{uav123} and LaSOT~\citep{fan2019lasot}, we utilize precision plot and success plot metrics in a one-pass evaluation (OPE) scenario. As for the VOT2018 challenge ~\citep{vot2018}, we introduce a reinitialization mechanism five frames after the tracker lost the target.\\
\textbf{Computing Infrastructures}. We conduct experiments on a computer with three Nvidia GeForce RTX 2080Ti and one Nvidia GeForce RTX 3090 GPUs, an Intel(R) Core(TM) i9-10900X CPU @ 3.70GHz, running Ubuntu 18.04.5 LTS.

\subsection{Implementation Details}
Our experiment is implemented in PyTorch. In momentum-based perturbation generation, maximum noise magnitude $\epsilon$ is 64, candidate number $C$ is 15, iteration number $k$ is 128, momentum factor $\mu$ is 0.5, trade-off factor $\iota$ is 0.4. Same to momentum generator, the texture-based generator produces adversarial sets with capacity $C$ as well.\par
To pretrain the Actor-Critic Network for key patch selection, we set PPO epoch, clipping parameter $\rho$, buffer capacity, and maximum gradient normalization to 10, 0.2, 500, and 0.5, respectively. As for patch number $\mathcal{P}$, we exploit the grid search strategy and set $\mathcal{P}$ as {2, 4, 8, 16, 32}. For balancing selection efficiency and final impact on tracking performance, $\mathcal{P}$ is parameterized to 16. \par
In the same way, the combination of ratio threshold $\tau_1$ and $\tau_2$ is set to 1.5 and 0.4. trade-off factor $\gamma$ is set to 0.4, video candidate number $n$ is naturally set to 20 out of 30, gradient candidate number $K$ is assigned to be 100, and the number of attack queries $\mathcal{N_A}$ can be 60. 

\subsection{Overall Attack Results}
\textbf{Results on VOT2018}. Table \ref{table_vot2018} compares the overall results of these trackers on the VOT2018 dataset. We exploit randomly generated noises as well as perturbations computed by IoU Attack~\citep{jia2021iou} and compare them with our proposed method. Specifically, our algorithm outperforms IoU Attack concerning accuracy in DaSiamRPN and ADNet by 8.45\% and 5.82\%, respectively. Furthermore, in terms of robustness, our approach exceeds IoU Attack in SiamPRN++, DasiamRPN, and ADNet by 9.32\%, 3.21\%, and 2.97\%. As for EAO (Expected Average Overlap) in SiamRPN++ and ADNet, we have achieved 6.2\% and 7.9\% improvement. \\
\textbf{Results on OTB100}. As shown in Figure \ref{otb100_img}, we draw success and precision plots of various trackers selected according to their categories and tested on OTB100. Compared to the original tracking performance, our black-box attack method can reduce the AUC score and visually change the curves' shape. Meanwhile, we correspondingly visualize the results of a white-box One-Shot Attack~\citep{chen2020one} and check the difference. Meanwhile, Table \ref{otb100_table} illustrates the success and precision rates of original videos, random perturbations, One-Shot Attack, IoU Attack, and our method.\\
\textbf{Results on UAV123 and LaSOT} Depicted in Figure \ref{uavlasot}, tracking results of different trackers are illustrated based on UAV123 and LaSOT. With our attack method, the AUC score of success plots tested on UAV123 are decreased by 4.3\%, 10.8\%, and 17.4\% for PrDiMP, SiamRPN++, and ADNet individually. In the meantime, the same score of success plots calculated on LaSOT are reduced by 6.6\%, 9.0\%, 22.5\%, and 11.8\% for PrDiMP, SiamRPN++, DaSiamRPN, and ADNet respectively.
\begin{table}[htbp]
\begin{center}
\begin{minipage}{\textwidth}
\caption{\label{vot}Attack results of SiamRPN++ ~\citep{siamrpn++}, DaSiamRPN~\citep{dasiamrpn}, PrDiMP ~\citep{prdimp}, ADNet~\citep{yun2017action}, and TrTr~\citep{zhao2021trtr} on VOT2018~\citep{vot2018}, evaluated using Accuracy, Robustness, and EAO(Expected Average Overlap).}
\label{table_vot2018}
\begin{tabular*}{\textwidth}{@{\extracolsep{\fill}}lcccc@{\extracolsep{\fill}}}
\toprule%
\multirow{2}{*}{Trackers}                  & \multicolumn{4}{c}{Accuracy$\uparrow$}     \\ \cmidrule{2-5}
& Original &  Random & IoU Attack & \textbf{Ours}\\
\midrule
SiamRPN++(R) & 60.30\%&59.12\%& \textbf{56.84}\% &57.01\%\\ 
DaSiamRPN & 58.52\%& 57.14\%& 53.19\%& \textbf{48.68}\%\\
PrDiMP50 & 61.80\%& 60.86\%& \textbf{57.29}\%& 58.12\%\\
ADNet & 50.80\%& 48.28\%& 39.53\%& \textbf{37.14}\%\\
TrTr & 60.65\%& 60.12\%& \textbf{57.88}\%& 58.84\%\\
\botrule
\end{tabular*}
\begin{tabular*}{\textwidth}{@{\extracolsep{\fill}}lcccc@{\extracolsep{\fill}}}
\toprule%
\multirow{2}{*}{Trackers}                  & \multicolumn{4}{c}{Robustness$\downarrow$}     \\ \cmidrule{2-5}
& Original &  Random & IoU Attack & \textbf{Ours}\\
\midrule
SiamRPN++(R) & 0.235 & 0.289 & 1.169 & \textbf{1.278}\\ 
DaSiamRPN & 0.276 & 0.295 & 1.214 & \textbf{1.253}\\
PrDiMP50 & 0.165 & 0.171 & \textbf{0.377} & 0.352\\
ADNet & 0.314 & 0.337 & 1.412 & \textbf{1.454}\\
TrTr & 0.110 & 0.121 & \textbf{0.227} & 0.193\\
\botrule
\end{tabular*}
\begin{tabular*}{\textwidth}{@{\extracolsep{\fill}}lcccc@{\extracolsep{\fill}}}
\toprule%
\multirow{2}{*}{Trackers}                  & \multicolumn{4}{c}{EAO(Expected Average Overlap)$\uparrow$}     \\ \cmidrule{2-5}
& Original &  Random & IoU Attack & \textbf{Ours}\\
\midrule
SiamRPN++(R) & 0.415 & 0.351 & 0.129 & \textbf{0.121}\\ 
DaSiamRPN & 0.382 & 0.347 & \textbf{0.124} & 0.159\\
PrDiMP50 & 0.442 & 0.425 & \textbf{0.275} & 0.311\\
ADNet & 0.329 & 0.317 & 0.113 & \textbf{0.104}\\
TrTr & 0.493 & 0.488 & \textbf{0.336} & 0.343\\
\botrule
\end{tabular*}
\end{minipage}
\end{center}
\end{table}
\begin{table}[htbp]
\begin{center}
\begin{minipage}{\textwidth}
\caption{Attack Results of SiamRPN++(ResNet50), SiamRPN++(Mobilev2), DaSiamRPN, ADNet, and TrTr on OTB100~\citep{otb}, evaluated using success rate and precision. As OPE(One Pass Evaluation) dataset, OTB100 can also be perturbed by white-box attacks, like One-Shot Attack~\citep{chen2020one}, which as it should be, outperforms black-box algorithms, and is highlighted in italic font.}
\label{otb100_table}
\renewcommand\arraystretch{1.2}
\begin{tabular*}{\textwidth}{@{\extracolsep{\fill}}lccccc@{\extracolsep{\fill}}}
\toprule%
\multirow{2}{*}{Trackers}                  & \multicolumn{5}{c}{Success Rate$\uparrow$}     \\ \cmidrule{2-6}
& Original & Random & IoU Attack & One-Shot Attack & \textbf{Ours}\\
\midrule
SiamRPN++(R)&69.64\% &65.21\% &49.58\% &\textit{25.22}\% &\textbf{48.09}\%\\
SiamRPN++(M) & 66.06\% &59.41\%  &\textbf{42.73}\%  &\textit{35.94}\% & 45.02\% \\
DaSiamRPN & 65.82\% &63.91\%  &\textbf{53.24}\%  &\textit{37.60}\% & 56.66\% \\
ADNet & 63.71\% &61.76\%  &53.80\%  &\textit{30.98}\% & \textbf{51.92}\% \\
TrTr & 71.53\% &68.32\%  &56.32\%  &\textit{41.88}\% & \textbf{54.16}\% \\
PrDiMP50 & 69.50\% &66.03\%  &46.54\%  &\textit{28.10}\% & \textbf{44.52}\% \\
\botrule
\end{tabular*}
\begin{tabular*}{\textwidth}{@{\extracolsep{\fill}}lccccc@{\extracolsep{\fill}}}
\toprule%
\multirow{2}{*}{Trackers}                  & \multicolumn{5}{c}{Precision$\uparrow$}     \\ \cmidrule{2-6}
& Original & Random & IoU Attack & One-Shot Attack & \textbf{Ours}\\
\midrule
SiamRPN++(R)&91.42\% &86.13\% &\textbf{63.19}\% &\textit{33.68}\% & 63.68\%\\
SiamRPN++(M) & 86.43\% &79.76\%  &62.18\%  &\textit{26.41}\% & \textbf{61.29}\% \\
DaSiamRPN & 86.50\% &81.23\%  &64.78\%  &\textit{29.65}\% & \textbf{63.09}\% \\
ADNet & 88.13\% &84.15\%  &\textbf{51.20}\%  &\textit{20.85}\% & 54.55\% \\
TrTr & 92.81\% &87.86\%  &68.74\%  &\textit{45.85}\% & \textbf{67.66}\% \\
PrDiMP50 & 89.73\% &87.24\%  &70.88\%  &\textit{38.10}\% & \textbf{69.96}\% \\
\botrule
\end{tabular*}
\end{minipage}
\end{center}
\end{table}
\subsection{Ablation Study of Key Patch Selection}
We conduct a series of experiments to evaluate the impact of the key patch selection module. Discrimination model PrDiMP is selected as our baseline and tracking results on VOT2018 are shown in Figure \ref{fig3}. As we can conclude from Figure \ref{fig3}, we query fewer times in black-box settings to reach a similar perturbation magnitude $\epsilon$ using Key Patch Selection. Meanwhile, the average IoU scores on unlost frames remain much smaller than DIMBA Attack without the Key Patch Selection module.
\subsection{Comparison with Previous Works}
According to our understanding, the overall computational complexity of IoU Attack~\citep{jia2021iou} is $\mathcal{O}(KNL)$, where $K$ is the number of epochs for choosing perturbations on each frame, $N$ is the candidate number of random noises, $L$ is the length of the video clip. Whereas in our algorithm, our query complexity can be reduced to $\mathcal{O}(KN+C)$, where $C$ is a constant number independent of $L$. The comparison in computational efficiency between IoU Attack and our approach is illustrated in Table \ref{querytimes}.
Furthermore, we also illustrate the comparison with the One-Shot Attack~\citep{chen2020one} in Table \ref{whitebox}.
\begin{minipage}{\textwidth}
\begin{minipage}[t]{0.5\textwidth}
\makeatletter\def\@captype{table}
\caption{Evolving success rate and precision based on perturbations within different scopes.}
\resizebox{\textwidth}{27mm}{
\begin{tabular}{ccc}
    \toprule
    PrDiMP50     & Success Rate     & Precision \\
    \midrule
     Original & 0.695  & 0.898    \\
    Random Noise     & 0.663 & 0.871      \\
    $\epsilon=16$     & 0.587      & 0.833  \\
    $\epsilon=24$     & 0.563       & 0.842 \\
    $\epsilon=32$     & 0.551       & 0.810 \\
    $\epsilon=40$     & 0.504       & 0.779 \\
    $\epsilon=48$     & 0.498       & \textbf{0.700} \\
    $\epsilon=56$     & 0.496      & 0.712 \\
    $\epsilon=64$     & 0.511       & 0.763 \\
    SSIM=0.92     & 0.522       & 0.821 \\
    SSIM=0.84    & \textbf{0.445}       & 0.809 \\
    SSIM=0.76     & 0.453   & 0.813\\
    \bottomrule
\end{tabular}}

\label{sample-table}
\end{minipage}
\begin{minipage}[t]{0.5\textwidth}
\makeatletter\def\@captype{table}
\caption{Comparison of average query times between IoU and DIMBA Attack using SiamPRN++(R).}
\label{querytimes}
\resizebox{\textwidth}{12mm}{
\begin{tabular}{ccc}
    \toprule
    Datasets    & IoU Attack     & DIMBA \\
    \midrule
    OTB100 & 81460  &  \textbf{43295}   \\
    LaSOT & 228570 & \textbf{186285}      \\
    VOT2018  &  \textbf{96890}      & 98792  \\
    UAV123  & 129802       & \textbf{108901}  \\

    \bottomrule
\end{tabular}
}
\makeatletter\def\@captype{table}
\vspace{1.0mm}
\caption{Evaluation on OTB100 between One-Shot Attack and DIMBA.}
\label{whitebox}
\resizebox{\textwidth}{12mm}{
\begin{tabular}{ccc}
    \toprule
    Trackers     & One-Shot    & DIMBA \\
    \midrule
    SiamPRN++$_{success}$ & \textit{0.252} & 0.481     \\
    SiamRPN++$_{precision}$ & \textit{0.337} & 0.636      \\
    SiamMask$_{success}$  &    \textit{0.481}    &  0.585 \\
    SiamMask$_{precision}$  &  \textit{0.650}      &  0.740\\

    \bottomrule
\end{tabular}
}
\end{minipage}
\end{minipage}
\vspace{0.8mm}

\section{Conclusions}

In this work, we propose an effective and efficient query-based black-box attack for SOT. An Actor-Critic key patch selection module is exploited to reduce redundant noises and increase query efficiency. Meanwhile, the combination of texture-based and momentum-based perturbation generators diverse potential adversarial directions and introduce heavily damaged tracking performance. Compared with existing works, our method requires fewer queries on SOT and less perturbation from the perspective of a whole video clip but maintains competitive, even better manipulating results. The experiments in both long-term and short-term datasets across three major categories of trackers demonstrate the effectiveness of our framework. We hope this work can elucidate the source of vulnerabilities in these trackers, optimistically paving the way for more powerful ones.

\end{document}